\documentclass[sigconf]{acmart}
\usepackage[utf8]{inputenc} 
\usepackage[T1]{fontenc}    
\usepackage{hyperref}       
\usepackage{url}            
\usepackage{booktabs}       
\usepackage{amsfonts}       
\usepackage{nicefrac}       
\usepackage{microtype}      
\usepackage{xcolor}         
\usepackage{graphicx}       
\usepackage{wrapfig}
\usepackage{enumitem}
\usepackage{multirow} 
\usepackage[linesnumbered,ruled,lined, noend]{algorithm2e}
\usepackage{subcaption}
\usepackage{pifont}

\newlength{\textfloatsepsave} 
\AtBeginDocument{%
  \providecommand\BibTeX{{%
    \normalfont B\kern-0.5em{\scshape i\kern-0.25em b}\kern-0.8em\TeX}}}

\setcopyright{none}
\renewcommand\footnotetextcopyrightpermission[1]{}
\acmConference[KDD '22 DSHealth Workshop]{}{August 14, 2022}{Washington DC}

\begin{document}

\title{Pseudo value-based Deep Neural Networks for Multi-state Survival Analysis}

\author{Md Mahmudur Rahman}
\email{mrahman6@umbc.edu}
\affiliation{%
  \institution{University of Maryland, Baltimore County}
  \city{Baltimore}
  \state{Maryland}
  \country{USA}
}

\author{Sanjay Purushotham}
\email{psanjay@umbc.edu}
\affiliation{%
  \institution{University of Maryland, Baltimore County}
  \city{Baltimore}
  \state{Maryland}
  \country{USA}
}

\renewcommand{\shortauthors}{Md Mahmudur Rahman and Sanjay Purushotham}

\begin{abstract}
Multi-state survival analysis (MSA) uses multi-state models for the analysis of time-to-event data. In medical applications, MSA can provide insights about the complex disease progression in patients. A key challenge in MSA is the accurate subject-specific prediction of multi-state model quantities such as transition probability and state occupation probability in the presence of censoring. Traditional multi-state methods such as Aalen-Johansen (AJ) estimators and Cox-based methods are respectively limited by Markov and proportional hazards assumptions and are infeasible for making subject-specific predictions. Neural ordinary differential equations for MSA relax these assumptions but are computationally expensive and do not directly model the transition probabilities. 
To address these limitations, we propose a new class of pseudo-value-based deep learning models for multi-state survival analysis, where we show that pseudo values - designed to handle censoring - can be a natural replacement for estimating the multi-state model quantities when derived from a consistent estimator. In particular, we provide an algorithm to derive pseudo values from consistent estimators to directly predict the multi-state survival quantities from the subject's covariates. Empirical results on synthetic and real-world datasets show that our proposed models achieve state-of-the-art results under various censoring settings.
\end{abstract}



\maketitle
\section{Introduction}
Multi-state survival analysis (MSA) is the problem of analyzing time-to-event data using multi-state models (MSM). Multi-state models~\cite{hougaard1999multi2} are models of a continuous-time stochastic process that capture the movement of subjects among a finite number of healthy and/or disease states. 
Thus, multi-state modeling can provide insights into the disease progression by providing a detailed view of disease or recovery trajectory in patients. This helps to predict the probability of future events after a given history and thus, can improve the clinician's decision-making ability for survival analysis. 
\begin{figure}[ht]
    \centerline{\includegraphics[width=0.42\textwidth]{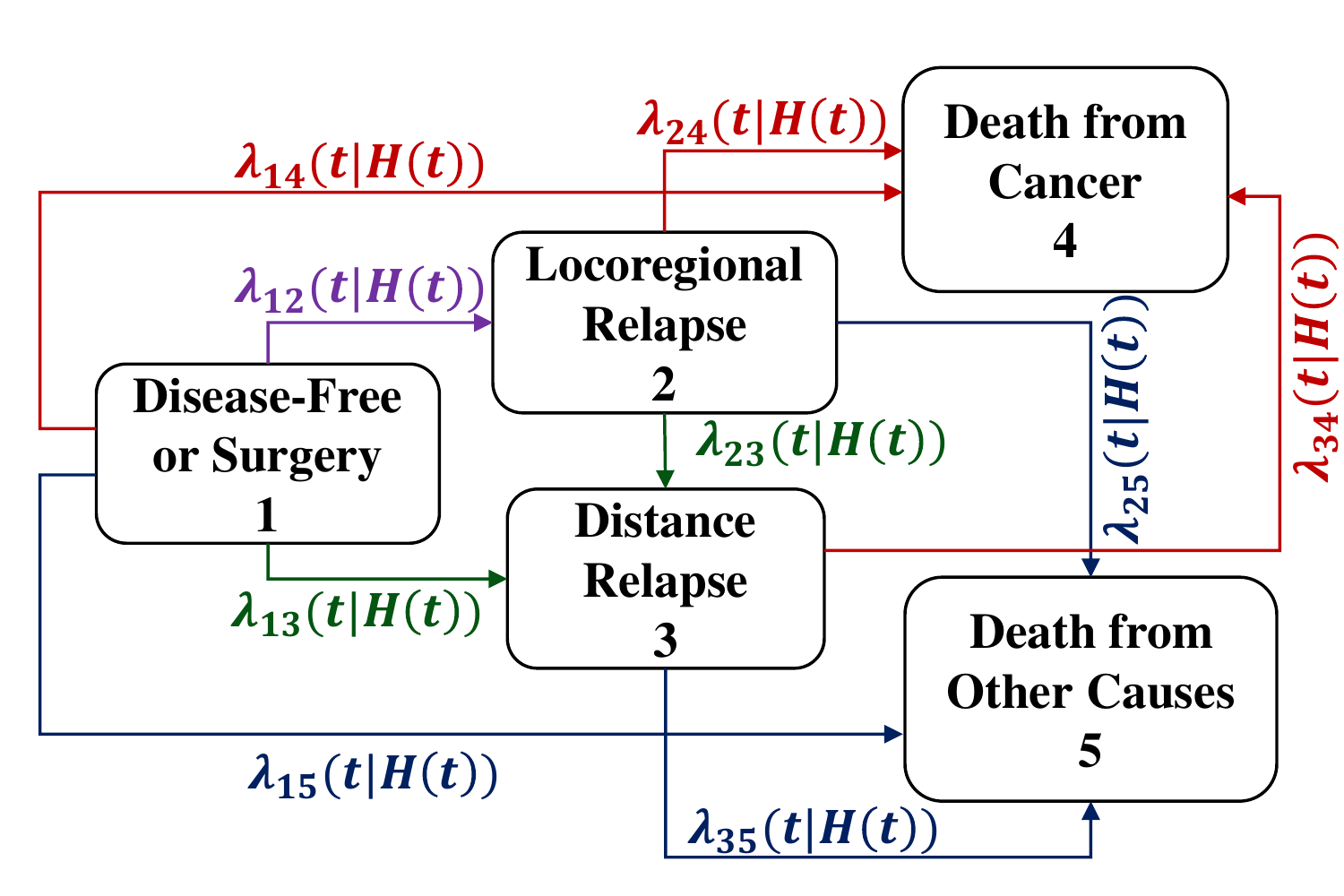}}
    \caption{\small Graphical representation of a breast-cancer patient's multi-state model, where rectangles represent states ($S=\{1, 2, ..., 5\})$ and arrows represent the instantaneous state-to-state transitions. A disease-free patient or a patient who had surgery can move to two clinically relevant intermediate states, i.e., locoregional relapse and distant relapse, until one of the two absorbing states (cancer-specific death and death by other causes) is observed.  $\lambda_{jk}(t|H(t))$ denotes the transition intensity functions. 
}
\label{fig:example}
\end{figure}
Figure \ref{fig:example} shows an example of a multi-state model for breast cancer progression. Here, a patient who is disease-free or had surgery can transition to locoregional relapse or distant relapse before reaching the death state. 
Multi-state survival analysis deals with the estimations of the multi-state quantities such as (a) \textbf{state occupation probability (SOP)} (the probability that a subject will be in a state $k$ at time $t$,), (b) the \textbf{transition probability} (the probability of transition to a state $k$ at time $t$ from another state $j$ at time $s$), and (c) \textbf{dynamic SOP} - which is the state occupation probabilities at some future time point $t$, given the event history (such as clinical information) is available up to a given time point $s$.
A variety of statistical and machine learning approaches have been developed over the years including non-parametric Aalen-Johansen (AJ) estimators~\cite{aalen1978empirical}, Cox-based
semi-parametric methods~\cite{de2010mstate}, parametric multi-state methods~\cite{jackson2016flexsurv}, and neural network-based methods -SurvNODE- \cite{groha2020neural} to estimate these multi-state quantities. 
Often, these approaches make strong assumptions such as linearity, proportional hazard, and Markov assumptions for each state or transition, which rarely hold in practice \citep{meira2009multi}. Moreover, many of these methods do not provide subject-specific predictions and do not handle the censoring well. 
Furthermore, these existing MSA methods cannot obtain subject-specific predictions of transition probability for non-Markov data because finding consistent estimators for non-Markov data has been understudied in the literature. 
Thus, new approaches that overcome these issues for multi-state survival analysis are in great demand.

\indent In this paper, we introduce pseudo value-based deep learning models for multi-state survival analysis, denoted as \textbf{msPseudo}, for estimating the multi-state quantities by treating the complex multi-state survival modeling as a regression analysis problem. \texttt{msPseudo} consists of a deep neural network that takes covariates as inputs to estimate a multi-state quantity (e.g., SOP) via a pseudo value regression task. 
\texttt{msPseudo} uses pseudo values as response variables because pseudo values have been shown to efficiently handle censoring for subject-specific survival predictions in survival analysis~\cite{zhao2020deep} and competing risks analysis~\cite{rahman2021deeppseudo}. 
Inspired by these works, we propose to use pseudo values as a replacement for multi-state quantities to handle censored observations. 
However, we cannot simply employ the estimators (Kaplan-Meier and Nelson–Aalen estimators) used in the earlier works~\cite{zhao2020deep,rahman2021deeppseudo} to derive pseudo values for multi-state quantities since these estimators are inconsistent especially for real-world non-Markov data and can result in large estimation errors. 
Therefore, we introduce a simple algorithm to derive the pseudo values from consistent estimators such as AJ and Landmark AJ~\cite{spitoni2018prediction} estimators by testing the Markovianity of the data using statistical significance tests - Commenges-Andersen (CA) test~\cite{commenges1995score} and log-rank statistic based tests~\cite{titman2020general}.  
Our algorithm provably obtains pseudo values from consistent estimators for both Markov and non-Markov data. 
Along with consistent pseudo-values, another advantage of our proposed model is that it does not make any underlying linear or proportional hazards assumptions and thus, can model the non-linear covariate effects during the prediction of subject-specific multi-state quantities. 
Therefore, our proposed {msPseudo} is simple yet flexible and overcomes the limitations of the existing multi-state survival models. 
We conducted extensive experiments on both simulated and real-world datasets to show that our proposed models achieve state-of-the-art performance in predicting multi-state survival quantities under various censoring settings. 

\vspace{-15pt}
\section{Our Proposed Pseudo value-based Deep Neural Networks}
We first describe the derivation of pseudo values for multi-state quantities before discussing our proposed pseudo value-based deep neural networks.

\textbf{Multi-state Survival Quantities:} A multi-state process is a continuous time stochastic process $\{X(t),t\in \mathcal{T}\}$, taking values in the (discrete-state) finite state space $S=\{1, 2, ..., K\}$, where $\mathcal{T}=[0,\tau],\tau<\infty$. MSA deals with the estimation of the multi-state quantities
\citep{spitoni2018prediction}: 
\textbf{Transition probability} is the probability of transition to a state k at time t from another state j at time s, defined as ${\mathcal{P}}_{jk}(s, t) = P(X(t) = k|X(s) = j, H(s)), \forall{j,k\in S}$
\textbf{State Occupation Probability (SOP)} is the probability that a subject will be in state k at time t, defined as,	$\pi_k(t)=P(X(t)=k); k\in S$; and 
\textbf{Dynamic SOP} is the SOP at some future time point t, given
the event history is available up to a given time point s, defined as $\pi_k(t|s)=P(X(t)=k|H(s))$.

\textbf{Pseudo values for multi-state quantities: } Multi-state survival datasets are subject to censoring, i.e., incomplete information about the stochastic process (for example, event or transition information missing due to loss to follow-up). Therefore, direct modeling of the event time or status with respect to covariates is challenging for censored observations. Inspired by the recent works~\cite{zhao2020deep,rahman2021deeppseudo}, we propose to use pseudo values as a substitute for the estimation of subject-specific multi-state model quantities in the presence of censoring. Thus, we estimate the pseudo values for a multi-state quantity of interest as $ \hat{y_i}(t^*)= n \hat{y}(t^*)-(n-1)\hat{y}^{-i}(t^*) $, where $\hat{y}(t^*)$ is an estimate of a consistent estimator, based on a $n$ samples, $\hat{y}^{-i}(t^*)$ is an estimate from the same estimator based on leave-one-out $(n-1)$ samples, obtained by omitting the $i^{th}$ subject. \\
\begin{algorithm}[tb]
 \caption{Pseudo value derivation algorithm}
 \SetCustomAlgoRuledWidth{0.45\textwidth}
\textbf{Inputs: }{Multi-state data, Selection of CA or log-rank test, $\epsilon$}\\
\textbf{Output: }{Pseudo values }\\
 \textbf{For SOP}:\\
  Choose either AJ or LMAJ estimator to derive pseudo values since both are consistent and same quantity~\citep{datta2001validity}.\\
 \textbf{For Dynamic SOP}:\\ 
 Perform CA test on the entire training data\\
 \eIf{P-value of the test is statistically significant for the violation of overall Markov assumption in the data}{
  Pseudo values $\leftarrow$ LMAJ estimator\;
  }
  {
  Pseudo values $\leftarrow$ AJ estimator\;
  }
\textbf{For TP}:\\ 
 Perform log rank test for checking transition-specific Markovianity.\\
 \eIf{P-value of the test is statistically significant for the violation of Markov assumption of a transition in the data\\}{
  \eIf{landmark population $< \epsilon$}{
     Pseudo values $\leftarrow$ AJ estimator\;
  }{
  Pseudo values $\leftarrow$ LMAJ estimator\;
  }
  }{
  Pseudo values $\leftarrow$ AJ estimator\;
  }
\end{algorithm}

\textbf{Consistent pseudo value derivation via Markov assumption testing:}
Pseudo values for MSA can be derived from an unbiased and consistent estimator such as the AJ estimator. Theoretical analysis of the consistency of the AJ and LMAJ estimators can be found in \cite{putter2018non}. However, the AJ estimator is inconsistent for non-Markov data and can result in large estimation errors  \citep{titman2020general}. 
Thus, recently, researchers have proposed Landmark AJ (LMAJ)~\cite{putter2018non} as a consistent and robust estimator for estimating the pseudo values for non-Markov data. However, the AJ estimator is known to be more efficient than LMAJ \citep{titman2020general} (when the Markov assumptions are valid), and in any practical scenario, the appropriateness of the Markov assumptions for a specific dataset remains unknown in advance, thus, making it infeasible to use just one estimator for pseudo value estimation.
To address this important challenge, we introduce and describe a pseudo value derivation algorithm shown in \textbf{Algorithm 1} to efficiently derive pseudo values by selecting consistent estimators by testing the underlying Markovian assumptions. 
Our algorithm takes as input the multi-state survival data and selects a consistent estimator to obtain pseudo values by testing the Markovianity assumptions in the dataset by using statistical significance tests such as Commenges-Andersen (CA) test \citep{commenges1995score} or log-rank statistic-based tests \citep{titman2020general}. 
CA and log-tank tests use a test statistic (usually $\chi^{2}$ statistic) and its corresponding p-values to identify the violation of Markov assumptions in the data. 
Note that $\epsilon$ is chosen based on the minimum size of the population in a landmark state. We fix $\epsilon = 1$ in our experiments.

\textbf{Proposed Model:}
We propose \textbf{msPseudo} - a first-of-its-kind, pseudo value-based deep
learning model for multi-state survival analysis. 
\texttt{msPseudo} is a simple feedforward deep neural network which performs regression analysis to
predict the multi-state quantities, such as state occupation probability (SOP), dynamic SOP, and transition probability (TP), using pseudo values as the response variables, given the covariates. 
\texttt{msPseudo} captures the complex non-linear hidden relationship between the patient's characteristics,
i.e., the baseline covariates and 
the multi-state model quantities.
For an input $n\times p$ matrix of $p$ baseline covariates with $n$ individuals input, \texttt{msPseudo} returns the predictions of a multi-state quantity (SOP, dynamic SOP, or TP). For a multi-state dataset with $K$ states, predicted output of SOP and dynamic SOP for a subject at a prespecified vector of $M$ time points $\mathbf{t}=\{\tau_1,\tau_2,...,\tau_M\}$ is a $K \times M$ matrix. 
For the TP prediction task, the output is a $Q \times M$ matrix, where $Q$ is the number of transitions.  
We used the mean squared error (the mean squared difference between pseudo values (ground truth) and the predicted multi-state quantity) as the loss function ($L$) to train our {msPseudo} model.

\begin{table*}[ht]
\centering
\caption{Comparison of SOP predictions on the Nonlinear Markov and Non-Markov datasets.}
\vspace{-12pt}
\label{tab:nonmarkov_nonlinear_SOP}
\renewcommand{\arraystretch}{1.1}
\resizebox{0.9\textwidth}{!}{
\begin{tabular}{c|cccccccc||cccccccccc}
\hline
\multirow{3}{*}{\large \textbf{Algorithm}} &
  \multicolumn{8}{c||}{\textbf{Nonlinear Markov}} &
  \multicolumn{10}{c}{\textbf{Nonlinear Non-Markov}} \\ \cline{2-19} 
 &
  \multicolumn{4}{c||}{\textbf{iAUC ($\uparrow$ better)}} &
  \multicolumn{4}{c||}{\textbf{iBS ($\downarrow$ better)}} &
  \multicolumn{5}{c||}{\textbf{iAUC ($\uparrow$ better)}} &
  \multicolumn{5}{c}{\textbf{iBS ($\downarrow$ better)}} \\ \cline{2-19} 
 &
  \multicolumn{1}{c|}{\textbf{S1}} &
  \multicolumn{1}{c|}{\textbf{S2}} &
  \multicolumn{1}{c||}{\textbf{S3}} &
  \multicolumn{1}{c||}{\textcolor{blue}{\textbf{Avg}}} &
  \multicolumn{1}{c|}{\textbf{S1}} &
  \multicolumn{1}{c|}{\textbf{S2}} &
  \multicolumn{1}{c||}{\textbf{S3}} &
  \multicolumn{1}{c||}{\textcolor{blue}{\textbf{Avg}}} &
  \multicolumn{1}{c|}{\textbf{S1}} &
  \multicolumn{1}{c|}{\textbf{S2}} &
  \multicolumn{1}{c|}{\textbf{S3}} &
  \multicolumn{1}{c||}{\textbf{S4}} &
  \multicolumn{1}{c||}{\textcolor{blue}{\textbf{Avg}}} &
  \multicolumn{1}{c|}{\textbf{S1}} &
  \multicolumn{1}{c|}{\textbf{S2}} &
  \multicolumn{1}{c|}{\textbf{S3}} &
  \multicolumn{1}{c||}{\textbf{S4}} &
  \textcolor{blue}{\textbf{Avg}} \\ \hline
\multicolumn{1}{c|}{\textbf{msCoxPH}} &
  \multicolumn{1}{c|}{0.94} &
  \multicolumn{1}{c|}{0.60} &
  \multicolumn{1}{c||}{0.91} &
  \multicolumn{1}{c||}{\textcolor{blue}{0.82}} &
  \multicolumn{1}{c|}{0.37} &
  \multicolumn{1}{c|}{0.18} &
  \multicolumn{1}{c||}{\textbf{0.08}} &
  \textcolor{blue}{0.21} &
  \multicolumn{1}{c|}{0.65} &
  \multicolumn{1}{c|}{0.70} &
  \multicolumn{1}{c|}{0.58} &
  \multicolumn{1}{c||}{0.76} &
  \multicolumn{1}{c||}{\textcolor{blue}{0.67}} &
  \multicolumn{1}{c|}{0.20} &
  \multicolumn{1}{c|}{0.12} &
  \multicolumn{1}{c|}{0.13} &
  \multicolumn{1}{c||}{0.01} &
  \textcolor{blue}{0.11} \\ \hline
\multicolumn{1}{c|}{\textbf{LinearPseudo}} &
  \multicolumn{1}{c|}{0.92} &
  \multicolumn{1}{c|}{0.59} &
  \multicolumn{1}{c||}{0.92} &
  \multicolumn{1}{c||}{\textcolor{blue}{0.81}} &
  \multicolumn{1}{c|}{0.10} &
  \multicolumn{1}{c|}{0.16} &
  \multicolumn{1}{c||}{0.19} &
  \textcolor{blue}{0.15} &
  \multicolumn{1}{c|}{0.85} &
  \multicolumn{1}{c|}{0.72} &
  \multicolumn{1}{c|}{0.88} &
  \multicolumn{1}{c||}{0.71} &
  \multicolumn{1}{c||}{\textcolor{blue}{0.79}} &
  \multicolumn{1}{c|}{0.13} &
  \multicolumn{1}{c|}{0.11} &
  \multicolumn{1}{c|}{0.08} &
  \multicolumn{1}{c||}{0.01} &
  \textcolor{blue}{0.08} \\ \hline
\multicolumn{1}{c|}{\textbf{SurvNODE}} &
  \multicolumn{1}{c|}{0.95} &
  \multicolumn{1}{c|}{0.67} &
  \multicolumn{1}{c||}{0.91} &
  \multicolumn{1}{c||}{\textcolor{blue}{0.84}} &
  \multicolumn{1}{c|}{0.16} &
  \multicolumn{1}{c|}{\textbf{0.15}} &
  \multicolumn{1}{c||}{0.09} &
  \textcolor{blue}{0.13} &
  \multicolumn{1}{c|}{0.83} &
  \multicolumn{1}{c|}{0.70} &
  \multicolumn{1}{c|}{0.76} &
  \multicolumn{1}{c||}{0.54} &
  \multicolumn{1}{c||}{\textcolor{blue}{0.71}} &
  \multicolumn{1}{c|}{0.29} &
  \multicolumn{1}{c|}{0.14} &
  \multicolumn{1}{c|}{0.16} &
  \multicolumn{1}{c||}{0.01} &
  \textcolor{blue}{0.15} \\ \hline
\multicolumn{1}{c|}{\textbf{msPseudo}} &
  \multicolumn{1}{c|}{\textbf{0.97}} &
  \multicolumn{1}{c|}{\textbf{0.67}} &
  \multicolumn{1}{c||}{\textbf{0.98}} &
  \multicolumn{1}{c||}{\textcolor{blue}{\textbf{0.87}}} &
  \multicolumn{1}{c|}{\textbf{0.06}} &
  \multicolumn{1}{c|}{0.16} &
  \multicolumn{1}{c||}{0.16} &
  \textcolor{blue}{\textbf{0.13}} &
  \multicolumn{1}{c|}{\textbf{0.86}} &
  \multicolumn{1}{c|}{\textbf{0.80}} &
  \multicolumn{1}{c|}{\textbf{0.90}} &
  \multicolumn{1}{c||}{\textbf{0.77}} &
  \multicolumn{1}{c||}{\textcolor{blue}{\textbf{0.83}}} &
  \multicolumn{1}{c|}{\textbf{0.12}} &
  \multicolumn{1}{c|}{\textbf{0.09}} &
  \multicolumn{1}{c|}{\textbf{0.07}} &
  \multicolumn{1}{c||}{\textbf{0.01}} &
  \textcolor{blue}{\textbf{0.07}} \\ \hline
\end{tabular}}
\vspace{-5pt}
\end{table*}

\begin{table*}[ht]
\centering
\caption{Comparison of the iBS ($\downarrow$ better) scores for METABRIC \& EBMT datasets.}
\vspace{-12pt}
\label{tab:real_SOP_Dynamic}
\renewcommand{\arraystretch}{1.0}
\resizebox{0.85\textwidth}{!}{
\begin{tabular}{c|c|ccccc||ccccccc}
\hline
\multirow{2}{*}{\textbf{\begin{tabular}[c]{@{}c@{}}Prediction \\ Task\end{tabular}}} &
  \multirow{2}{*}{\textbf{Model}} &
  \multicolumn{5}{c||}{\textbf{METABRIC}} &
  \multicolumn{7}{c}{\textbf{EBMT}} \\ \cline{3-14} 
 &
   &
  \textbf{State 1} &
  \textbf{State 2} &
  \textbf{State 3} &
  \multicolumn{1}{c||}{\textbf{State 4}} &
  \textcolor{blue}{\textbf{Avg}} &
  \textbf{State 1} &
  \textbf{State 2} &
  \textbf{State 3} &
  \textbf{State 4} &
  \textbf{State 5} &
  \multicolumn{1}{c||}{\textbf{State 6}} &
  \textcolor{blue}{\textbf{Avg}} \\ \hline
\multirow{3}{*}{\textbf{\begin{tabular}[c]{@{}c@{}}State \\ Occupation \\ Probability\end{tabular}}} &
  \textbf{msCox} &
  0.33 &
  0.03 &
  0.04 &
\multicolumn{1}{c||}{0.24} &  
  \textcolor{blue}{0.16} &
  0.13 &
  0.14 &
  0.11 &
  0.15 &
  \textbf{0.003} &
\multicolumn{1}{c||}{0.15} &   
  \textcolor{blue}{0.12}\\
 &
  \textbf{SurvNODE} &
  0.30 &
  0.03 &
  0.04 &
\multicolumn{1}{c||}{0.21} &   
  \textcolor{blue}{0.14} &
  0.34 &
  0.14 &
  0.11 &
  0.17 &
  0.003 &
\multicolumn{1}{c||}{0.17} &    
  \textcolor{blue}{0.16}\\
 &
  \textbf{msPseudo} &
  \textbf{0.21} &
  \textbf{0.03} &
  \textbf{0.04} &
\multicolumn{1}{c||}{0.17} &   
  \textcolor{blue}{\textbf{0.11}} &
  \textbf{0.13} &
  \textbf{0.14} &
  \textbf{0.10} &
  \textbf{0.15} &
  0.01 &
\multicolumn{1}{c||}{\textbf{0.15}} &    
  \textcolor{blue}{\textbf{0.11}}\\ \hline \hline
\multirow{3}{*}{\textbf{\begin{tabular}[c]{@{}c@{}}Dynamic \\ SOP\\ Prediction\end{tabular}}} &
  \textbf{msCox} &
  0.32 &
  0.03 &
  0.03 &
\multicolumn{1}{c||}{0.23} &   
  \textcolor{blue}{0.15} &
  \textbf{0.10} &
  \textbf{0.11} &
  \textbf{0.08} &
  \textbf{0.12} &
  0.01 &
\multicolumn{1}{c||}{\textbf{0.14}} &   
  \textcolor{blue}{\textbf{0.09}}\\
 &
  \textbf{SurvNODE} &
  0.29 &
  0.03 &
  0.03 &
\multicolumn{1}{c||}{0.20} &   
  \textcolor{blue}{0.14} &
  0.13 &
  0.12 &
  0.11 &
  0.14 &
  \textbf{0.001} &
\multicolumn{1}{c||}{0.16} &   
  \textcolor{blue}{0.11}\\
 &
  \textbf{msPseudo} &
  \textbf{0.20} &
  \textbf{0.03} &
  \textbf{0.03} &
\multicolumn{1}{c||}{\textbf{0.17}} &   
  \textcolor{blue}{\textbf{0.11}} &
  0.11 &
  0.12 &
  0.09 &
  {0.13}&
  0.02 &
\multicolumn{1}{c||}{\textbf{0.14}} &  
  \textcolor{blue}{0.10}\\ \hline
\end{tabular}}
\vspace{-5pt}
\end{table*}

\begin{table*}[ht]
\caption{Comparison of the iBS ($\downarrow$ better) for the TP predictions of the models on METABRIC \& EBMT datasets.} 
\vspace{-12pt}
\label{tab:transition_real}
\renewcommand{\arraystretch}{1.1}
\resizebox{\textwidth}{!}{
\begin{tabular}{c|ccccccc||ccccccccccccc}
\hline
\multirow{2}{*}{\large\textbf{Model}} &
  \multicolumn{7}{c||}{\textbf{METABRIC}} &
  \multicolumn{13}{c}{\textbf{EBMT}} \\ \cline{2-21} 
 &
  \multicolumn{1}{c|}{\textbf{1$\rightarrow$2}} &
  \multicolumn{1}{c|}{\textbf{1$\rightarrow$3}} &
  \multicolumn{1}{c|}{\textbf{1$\rightarrow$4}} &
  \multicolumn{1}{c|}{\textbf{2$\rightarrow$3}} &
  \multicolumn{1}{c|}{\textbf{2$\rightarrow$4}} &
  \multicolumn{1}{c||}{\textbf{3$\rightarrow$4}} &
  \textcolor{blue}{\textbf{Avg}} &
  \multicolumn{1}{|c|}{\textbf{1$\rightarrow$2}} &
  \multicolumn{1}{c|}{\textbf{1$\rightarrow$3}} &
  \multicolumn{1}{c|}{\textbf{1$\rightarrow$5}} &
  \multicolumn{1}{c|}{\textbf{1$\rightarrow$6}} &
  \multicolumn{1}{c|}{\textbf{2$\rightarrow$4}} &
  \multicolumn{1}{c|}{\textbf{2$\rightarrow$5}} &
  \multicolumn{1}{c|}{\textbf{2$\rightarrow$6}} &
  \multicolumn{1}{c|}{\textbf{3$\rightarrow$4}} &
  \multicolumn{1}{c|}{\textbf{3$\rightarrow$5}} &
  \multicolumn{1}{c|}{\textbf{3$\rightarrow$6}} &
  \multicolumn{1}{c|}{\textbf{4$\rightarrow$5}} &
  \multicolumn{1}{c||}{\textbf{4$\rightarrow$6}} &
  \textcolor{blue}{\textbf{Avg}} \\ \hline
\textbf{msCox} &
  \multicolumn{1}{c|}{0.02} &
  \multicolumn{1}{c|}{0.03} &
  \multicolumn{1}{c|}{0.30} &
  \multicolumn{1}{c|}{0.05} &
  \multicolumn{1}{c|}{\textbf{0.11}} &
  \multicolumn{1}{c||}{\textbf{0.30}} &
  \textcolor{blue}{\textbf{0.14}} &
  \multicolumn{1}{c|}{0.07} &
  \multicolumn{1}{c|}{0.02} &
  \multicolumn{1}{c|}{\textbf{0.01}} &
  \multicolumn{1}{c|}{0.16} &
  \multicolumn{1}{c|}{0.05} &
  \multicolumn{1}{c|}{\textbf{0.01}} &
  \multicolumn{1}{c|}{0.09} &
  \multicolumn{1}{c|}{{0.12}} &
  \multicolumn{1}{c|}{\textbf{0.002}} &
  \multicolumn{1}{c|}{\textbf{0.12}} &
  \multicolumn{1}{c|}{\textbf{0.004}} &
  \multicolumn{1}{c||}{\textbf{0.06}} &
  \textcolor{blue}{{0.06}} \\ \hline
\textbf{msWeibull} &
  \multicolumn{1}{c|}{0.02} &
  \multicolumn{1}{c|}{0.03} &
  \multicolumn{1}{c|}{\textbf{0.17}} &
  \multicolumn{1}{c|}{0.10} &
  \multicolumn{1}{c|}{0.37} &
  \multicolumn{1}{c||}{0.45} &
  \textcolor{blue}{0.19} &
  \multicolumn{1}{c|}{0.05} &
  \multicolumn{1}{c|}{0.03} &
  \multicolumn{1}{c|}{0.09} &
  \multicolumn{1}{c|}{0.16} &
  \multicolumn{1}{c|}{0.05} &
  \multicolumn{1}{c|}{0.09} &
  \multicolumn{1}{c|}{0.13} &
  \multicolumn{1}{c|}{\textbf{0.07}} &
  \multicolumn{1}{c|}{0.11} &
  \multicolumn{1}{c|}{0.14} &
  \multicolumn{1}{c|}{0.10} &
  \multicolumn{1}{c||}{0.12} &
  \textcolor{blue}{0.09} \\ \hline
\textbf{msPseudo} &
  \multicolumn{1}{c|}{\textbf{0.02}} &
  \multicolumn{1}{c|}{\textbf{0.03}} &
  \multicolumn{1}{c|}{0.18} &
  \multicolumn{1}{c|}{\textbf{0.01}} &
  \multicolumn{1}{c|}{0.56} &
  \multicolumn{1}{c||}{0.79} &
  \textcolor{blue}{0.27} &
  \multicolumn{1}{c|}{\textbf{0.05}} &
  \multicolumn{1}{c|}{\textbf{0.02}} &
  \multicolumn{1}{c|}{0.02} &
  \multicolumn{1}{c|}{\textbf{0.16}} &
  \multicolumn{1}{c|}{\textbf{0.04}} &
  \multicolumn{1}{c|}{0.03} &
  \multicolumn{1}{c|}{\textbf{0.09}} &
  \multicolumn{1}{c|}{0.09} &
  \multicolumn{1}{c|}{0.01} &
  \multicolumn{1}{c|}{0.14} &
  \multicolumn{1}{c|}{0.02} &
  \multicolumn{1}{c||}{0.07} &
  \textcolor{blue}{\textbf{0.06}} \\ \hline
\end{tabular}}
\vspace{-5pt}
\end{table*}

\begin{table}[ht]
\centering
\vspace{-2pt}
\caption{Comparison of the iBS ($\downarrow$ better) scores for the Linear Non-Markov dataset.}
\vspace{-12pt}
\label{tab:NonMarkov_Linear}
\renewcommand{\arraystretch}{1.3}
\resizebox{0.49\textwidth}{!}{
\begin{tabular}{c|ccccccccccl}
\cline{1-11}
\multirow{2}{*}{\textbf{Model}} &
  \multicolumn{5}{c|}{\textbf{SOP}} &
  \multicolumn{5}{c}{\textbf{Dynamic SOP (s=1 year)}} &
   \\ \cline{2-11}
 &
  \textbf{S1} &
  \textbf{S2} &
  \textbf{S3} &
  \textbf{S4} &
  \multicolumn{1}{|c|}{\textcolor{blue}{\textbf{Avg}}} &
  \textbf{S1} &
  \textbf{S2} &
  \textbf{S3} &
  \textbf{S4} &
  \multicolumn{1}{|c}{\textcolor{blue}{\textbf{Avg}}} &
   \\ \cline{1-11}
\textbf{AJ} &
  0.21  &
  0.16  &
  \textbf{0.11}  &
  0.01  &
   \multicolumn{1}{|c|}{\textcolor{blue}{0.12}}  &
  0.11  &
  0.08  &
  0.05  &
  0.01 &
   \multicolumn{1}{|c}{\textcolor{blue}{0.06}} \\ 
\textbf{LMAJ} &
  0.21  &
  0.16  &
  0.11  &
  0.01  &
   \multicolumn{1}{|c|}{\textcolor{blue}{0.12}}  &
  0.15  &
  0.10  &
  0.06  &
  0.01  &
   \multicolumn{1}{|c}{\textcolor{blue}{0.08}} \\    
\textbf{msCox} &
  0.20  &
  0.15  &
  0.12  &
  0.01 &
  \multicolumn{1}{|c|}{\textcolor{blue}{0.12}} &
  0.11  &
  0.09  &
  0.05  &
  0.01  &
   \multicolumn{1}{|c}{\textcolor{blue}{0.06}} \\ 
\textbf{SurvNODE} &
  0.33  &
  0.20  &
  0.14  &
  0.01 &
  \multicolumn{1}{|c|}{\textcolor{blue}{0.17}} &
  0.14  &
  0.09  &
  0.05  &
  0.01  &
   \multicolumn{1}{|c}{\textcolor{blue}{0.07}} \\ 
\textbf{msPseudo} &
  \textbf{0.19 } &
  \textbf{0.15 } &
  {0.12 } &
  \textbf{0.01 } &
  \multicolumn{1}{|c|}{\textcolor{blue}{\textbf{0.12}}} &
  \textbf{0.04 } &
  \textbf{0.03 } &
  \textbf{0.02 } &
  \textbf{0.002 } &
   \multicolumn{1}{|c}{\textcolor{blue}{\textbf{0.02 }}}   
   \\ \cline{1-11}
\multirow{2}{*}{\textbf{Model}} &
  \multicolumn{10}{c}{\textbf{TP (s=1 year)}} \\ \cline{2-11} 
 &
  \textbf{1$\rightarrow$2} &
  \textbf{1$\rightarrow$3} &
  \textbf{1$\rightarrow$4} &
  \textbf{2$\rightarrow$1} &
  \textbf{2$\rightarrow$3} &
  \textbf{2$\rightarrow$4} &
  \textbf{3$\rightarrow$1} &
  \textbf{3$\rightarrow$2} &
  \textbf{3$\rightarrow$4} &
  \multicolumn{1}{|c}{\textcolor{blue}{\textbf{Avg}}} \\ \cline{1-11}
 \textbf{AJ}        & 0.05  & 0.04  & 0.02  & 0.11  & 0.03  & 0.01  & 0.05  & 0.06  & 0.01 & \multicolumn{1}{|c}{\textcolor{blue}{0.04}}  \\ 
\textbf{LMAJ}      & 0.05  & {0.04}  & \textbf{0.001}  & 0.10  & 0.04  & 0.01  & \textbf{0.02}  & 0.10  & 0.01 & \multicolumn{1}{|c}{\textcolor{blue}{0.04}}  \\ 
\textbf{msCox} &
  {0.05} &
  0.04 &
  0.02  &
  0.12  &
  {0.03} &
  0.01  &
  0.03  &
  0.04  &
  \textbf{0.003} &
  \multicolumn{1}{|c}{\textcolor{blue}{0.04}} \\
\textbf{msWeibull} &
  0.06  &
  0.04  &
  0.14  &
  \textbf{0.05} &
  0.03  &
  0.08  &
  {0.03} &
  0.05  &
  0.18 &
  \multicolumn{1}{|c}{\textcolor{blue}{0.07}}\\
\textbf{msPseudo} &
  \textbf{0.05}  &
  \textbf{0.04}  &
  {0.02} &
  0.13  &
  \textbf{0.03}  &
  \textbf{0.01} &
  0.05  &
  \textbf{0.03} &
  0.01 &
  \multicolumn{1}{|c}{\textcolor{blue}{\textbf{0.04}}} \\ \cline{1-11}
\end{tabular}}
\end{table}

\begin{figure*}[h]
\vspace{-3pt}
\centering
    \includegraphics[trim=0 147 0 150,clip,width=0.95\textwidth]{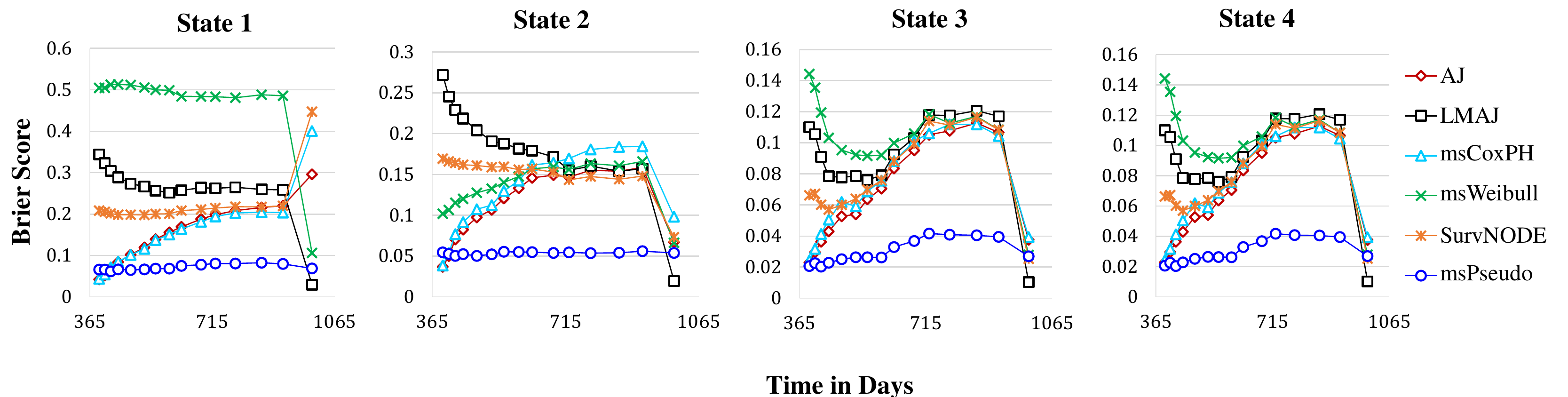}
        \vspace{-10pt}
    \caption{Comparison of time-dependent Brier score for the dynamic SOP prediction at s=1 year for \textbf{reversible Linear Non-Markov} dataset. Blue-circle represents our proposed \textbf{msPseudo} model.}
    \label{fig:Dynamic_Non_Markov_Linear}
    \vspace{-10pt}
\end{figure*}

\begin{table}[h]
\centering
\vspace{-2pt}
\caption{Comparison of the iAUC on high censoring settings (75\%) for SOP prediction on the Nonlinear Markov dataset. 
}
\vspace{-10pt}
\label{tab:Censoring_AUC}
\renewcommand{\arraystretch}{1.3}
\resizebox{0.47\textwidth}{!}{
\begin{tabular}{c|cccc||cccc}
\hline
\multirow{2}{*}{\textbf{Algorithm}} & \multicolumn{4}{c||}{\textbf{Incremental Censoring}}                    & \multicolumn{4}{c}{\textbf{Incremental Censoring}}                     \\ \cline{2-9} 
                           & \multicolumn{1}{c|}{\textbf{S1}}    & \multicolumn{1}{c|}{\textbf{S2}}    &
                           \multicolumn{1}{c||}{\textbf{S3}}    &
                           \textcolor{blue}{\textcolor{blue}{\textbf{Avg}}}    & \multicolumn{1}{c|}{\textbf{S1}}    & \multicolumn{1}{c|}{\textbf{S2}}    &
                           \multicolumn{1}{c||}{\textbf{S3}}    &
                           \textcolor{blue}{\textcolor{blue}{\textbf{Avg}}}    \\ \hline
\textbf{msCox}                      & \multicolumn{1}{c|}{0.54} & \multicolumn{1}{c|}{0.54} & \multicolumn{1}{c||}{0.52} & \textcolor{blue}{0.53} & \multicolumn{1}{c|}{0.52} & \multicolumn{1}{c|}{0.52} &
\multicolumn{1}{c||}{0.53} & \textcolor{blue}{0.52} \\ \hline
\textbf{msWeibull}    & \multicolumn{1}{c|}{0.51} & \multicolumn{1}{c|}{0.51} &
\multicolumn{1}{c||}{0.52} &
\textcolor{blue}{0.51} & \multicolumn{1}{c|}{0.52} & \multicolumn{1}{c|}{0.53} &
\multicolumn{1}{c||}{0.52} &
\textcolor{blue}{0.52} \\ \hline
\textbf{LinearPseudo} & \multicolumn{1}{c|}{0.52} & \multicolumn{1}{c|}{0.53} &
\multicolumn{1}{c||}{0.52} &
\textcolor{blue}{0.52} & \multicolumn{1}{c|}{0.51} & \multicolumn{1}{c|}{0.52} & \multicolumn{1}{c||}{0.53} & \textcolor{blue}{0.52} \\ \hline
\textbf{SurvNODE}                   & \multicolumn{1}{c|}{0.66} & \multicolumn{1}{c|}{0.61} &
\multicolumn{1}{c||}{0.65} & \textcolor{blue}{0.64} & \multicolumn{1}{c|}{0.69} & \multicolumn{1}{c|}{0.62} & \multicolumn{1}{c||}{0.64} & \textcolor{blue}{0.65} \\ \hline
\textbf{msPseudo}                   & \multicolumn{1}{c|}{\textbf{0.90}} & \multicolumn{1}{c|}{\textbf{0.65}} & \multicolumn{1}{c||}{\textbf{0.74}} & \textcolor{blue}{\textbf{0.77}} & \multicolumn{1}{c|}{\textbf{0.88}} & \multicolumn{1}{c|}{\textbf{0.63}} & \multicolumn{1}{c||}{\textbf{0.66}} & \textcolor{blue}{\textbf{0.72}} \\ \hline
\end{tabular}}
\end{table}

\vspace{-10pt}
\section{Experiments}
We conducted experiments on both simulated and real-world datasets to answer the following questions: (a) How do our proposed models compare against the existing MSA approaches for predicting multi-state quantities? (b) How well do our proposed models perform under a variety of censoring settings compared to other models?

\textbf{Simulation datasets:} We generated the following four simulation datasets (two Markov and two non-Markov datasets): \textit{(1) Time homogeneous Linear Markov Data; (2) Time homogeneous Nonlinear Markov Data; 
(3) Linear Reversible Non-Markov Data; (4) Nonlinear Reversible Non-Markov Data} - to obtain different datasets with varying Markov and linearity assumptions. 
For each dataset, we simulated 5000 examples with multiple transitions. The Markov datasets have three states and allow only forward transitions. The Non-Markov multi-state datasets consist of four states: states 1–3 are intermediate and interconnected states, and state 4 is an absorbing state, and they allow reverse transitions \citep{hoff2019landmark}. 
 
\textbf{Real-world datasets:} We used the following publicly available datasets for our experiments: (1) \textit{METABRIC~\cite{rueda2019dynamics} }
dataset contains 1975 breast cancer patient data with multiple transitions and 20 covariates collected over a 360-month study.
This multi-state dataset has four states: Surgery, Locoregional Relapse, Distance Relapse, and Death.  
(2)
\textit{EBMT}~\citep{de2011mstate} dataset contains 2279 transplantation patient data collected between 1985 and 1998. 
In this dataset, an alive patient in remission without recovery or adverse event can move to three possible distinct intermediate states, i.e., recovery, adverse event, and co-occurrence of recovery and adverse event, until one of the two absorbing states (death and relapse) is observed. 

\textbf{Censoring settings:} We investigate the impact of the higher rate of censoring (75\%) on MSA model performances under two censoring settings: incremental censoring (adding censored observations to a fixed number of uncensored observations) and induced censoring (inducing censored observations by flipping the label of transition status of the uncensored observations) settings \citep{rahman2021deeppseudo}.

\textbf{Prediction tasks:}  Given the covariates, we perform regression for estimating the multi-state quantities such as SOP, dynamic SOP, and TP. 
We compare the performances of the following multi-state models for these prediction tasks: 
\textbf{Non-parametric models:} AJ estimator (\texttt{AJ}) \citep{aalen2008survival}, LMAJ estimator (\texttt{LMAJ}) \citep{putter2018non}; 
\textbf{Parametric models:} Weibull parametric model (\texttt{msWeibull}) \citep{jackson2016flexsurv}, linear Pseudo value model (\texttt{LinearPseudo}) \cite{andersen2007regression}; \textbf{Semi-parametric model:} Multi-state Cox proportional hazard model 
(\texttt{msCox}) \citep{de2010mstate}; \textbf{Deep learning multi-state model:} \texttt{SurvNODE} \citep{groha2020neural}; \textbf{Our proposed model:} \texttt{msPseudo}

\textbf{Evaluations}: We evaluate the models in terms of \textbf{integrated Brier score (iBS)} \cite{spitoni2018prediction} and \textbf{integrated AUC (iAUC)} \cite{fawcett2006introduction}. 
We perform 5 runs of 5-fold cross-validation and report the average of these evaluation metrics. We train our models using Adam optimizer~\cite{kingma2014adam} to 10000 epochs with an early stopping criterion. Hyperparameter tuning (over batch size, learning rate, drop-out, number of layers, etc.) is performed to choose the best performing deep learning models.
A sigmoid activation function is used in the output layer to obtain the multi-state quantities from the predicted pseudo values.

\vspace{-5pt}
\section{Results and Discussion}
\textbf{Simulated data:}
Table \ref{tab:nonmarkov_nonlinear_SOP} shows that our \texttt{msPseudo} performs significantly better than \texttt{msCox} and \texttt{SurvNODE} for SOP prediction in Non-linear Non-Markov dataset in terms of iAUC and iBS metrics. This shows that our models work well with non-Markov assumptions and can capture non-linearity in the data. 
Our \texttt{msPseudo} outperforms other models in terms of iAUC and is comparable in terms of iBS for the Non-linear Markov dataset.
Table~\ref{tab:NonMarkov_Linear} shows that our model \texttt{msPseudo} performs significantly better than \texttt{msCox} and \texttt{SurvNODE} in SOP and Dynamic SOP prediction task on reversible Linear Non-Markov data. 
In the TP prediction task, \texttt{msPseudo} achieves similar or better results on 7 out of 9 transitions compared to the other models. 
We also graphically show the time-dependent Brier Score comparison in Figure \ref{fig:Dynamic_Non_Markov_Linear} for dynamic SOP prediction on Linear Non-Markov data. This figure demonstrates that \texttt{msPseudo} achieves $\sim$10\% improvement over other multi-state models.

\textbf{Real-world data:} The predictive performances for the real-world clinical data: METABRIC and EBMT are shown in Table \ref{tab:real_SOP_Dynamic} \& \ref{tab:transition_real}. Our model, \texttt{msPseudo}, outperforms all other models on METABRIC data for both SOP and Dynamic SOP prediction tasks. \texttt{msPseudo} also obtains lowest iBS for SOP prediction task on EBMT dataset while \texttt{msCox} performs similar or marginally better for Dynamic SOP predictions. 
In table \ref{tab:transition_real}, we observe that \texttt{msPseudo} gives overall better iBS performance, i.e., better prediction on both METABRIC and EBMT datasets. In some cases, our models give comparable performance to the msCox model due to the absence of covariates interaction effect, negligible violation of the proportional hazards and Markov assumptions. However, when averaged over all states and transitions (shown as the column \textcolor{blue}{Avg} in all tables), we find that our proposed models outperform msCox and other MSA methods.

\textbf{Various Censoring Settings:} 
Table~\ref{tab:Censoring_AUC} shows the iAUC results of different multi-state models on incremental and induced censoring settings for SOP prediction in the time-homogeneous nonlinear Markov dataset. 
From this table, we see that our \texttt{msPseudo} performs significantly better (more efficient in handling censoring using pseudo values) than other models in both incremental and induced censoring settings with a high censoring rate (75\% censoring).

\vspace{-5pt}
\section{Conclusion}
Multi-state survival analysis (MSA) is an important yet under-studied problem in time-to-event literature. Finding consistent estimators for non-Markov data is still an open problem in this field.  
In this paper, we proposed a first-of-its-kind novel pseudo value-based deep learning model, \textbf{msPseudo}, 
for estimating multi-state survival quantities in the presence of censoring without making any assumption on the underlying multi-state processes. We show that pseudo values can be a replacement for the estimation of multi-state quantities when derived from a consistent estimator. Through empirical experiments on simulated and real datasets, we demonstrated that our proposed models outperform other multi-state survival models under various censoring settings and for both Markov and non-Markov datasets. 
We believe this work lays the foundation for future investigations on the use of deep models for MSA, including explaining survival predictions and state-specific transition probabilities in real-world datasets.

\section*{Acknowledgement}
This work is supported by grant IIS–1948399 from the US National Science Foundation (NSF).

\bibliographystyle{ACM-Reference-Format}
\bibliography{mybib}


\begin{thebibliography}{20}


\ifx \showCODEN    \undefined \def \showCODEN     #1{\unskip}     \fi
\ifx \showDOI      \undefined \def \showDOI       #1{#1}\fi
\ifx \showISBNx    \undefined \def \showISBNx     #1{\unskip}     \fi
\ifx \showISBNxiii \undefined \def \showISBNxiii  #1{\unskip}     \fi
\ifx \showISSN     \undefined \def \showISSN      #1{\unskip}     \fi
\ifx \showLCCN     \undefined \def \showLCCN      #1{\unskip}     \fi
\ifx \shownote     \undefined \def \shownote      #1{#1}          \fi
\ifx \showarticletitle \undefined \def \showarticletitle #1{#1}   \fi
\ifx \showURL      \undefined \def \showURL       {\relax}        \fi
\providecommand\bibfield[2]{#2}
\providecommand\bibinfo[2]{#2}
\providecommand\natexlab[1]{#1}
\providecommand\showeprint[2][]{arXiv:#2}

\bibitem[Aalen et~al\mbox{.}(2008)]%
        {aalen2008survival}
\bibfield{author}{\bibinfo{person}{Odd Aalen}, \bibinfo{person}{Ornulf Borgan},
  {and} \bibinfo{person}{Hakon Gjessing}.} \bibinfo{year}{2008}\natexlab{}.
\newblock \bibinfo{booktitle}{\emph{Survival and event history analysis: a
  process point of view}}.
\newblock \bibinfo{publisher}{Springer Science \& Business Media}.
\newblock


\bibitem[Aalen and Johansen(1978)]%
        {aalen1978empirical}
\bibfield{author}{\bibinfo{person}{Odd~O Aalen} {and} \bibinfo{person}{S{\o}ren
  Johansen}.} \bibinfo{year}{1978}\natexlab{}.
\newblock \showarticletitle{An empirical transition matrix for non-homogeneous
  Markov chains based on censored observations}.
\newblock \bibinfo{journal}{\emph{Scandinavian Journal of Statistics}}
  (\bibinfo{year}{1978}), \bibinfo{pages}{141--150}.
\newblock


\bibitem[Andersen and Klein(2007)]%
        {andersen2007regression}
\bibfield{author}{\bibinfo{person}{Per~K Andersen} {and}
  \bibinfo{person}{John~P Klein}.} \bibinfo{year}{2007}\natexlab{}.
\newblock \showarticletitle{Regression analysis for multistate models based on
  a pseudo-value approach, with applications to bone marrow transplantation
  studies}.
\newblock \bibinfo{journal}{\emph{Scandinavian Journal of Statistics}}
  \bibinfo{volume}{34}, \bibinfo{number}{1} (\bibinfo{year}{2007}),
  \bibinfo{pages}{3--16}.
\newblock


\bibitem[Commenges and Andersen(1995)]%
        {commenges1995score}
\bibfield{author}{\bibinfo{person}{Daniel Commenges} {and}
  \bibinfo{person}{Per~Kragh Andersen}.} \bibinfo{year}{1995}\natexlab{}.
\newblock \showarticletitle{Score test of homogeneity for survival data}.
\newblock \bibinfo{journal}{\emph{Lifetime data analysis}} \bibinfo{volume}{1},
  \bibinfo{number}{2} (\bibinfo{year}{1995}), \bibinfo{pages}{145--156}.
\newblock


\bibitem[Datta and Satten(2001)]%
        {datta2001validity}
\bibfield{author}{\bibinfo{person}{Somnath Datta} {and} \bibinfo{person}{Glen~A
  Satten}.} \bibinfo{year}{2001}\natexlab{}.
\newblock \showarticletitle{Validity of the Aalen--Johansen estimators of stage
  occupation probabilities and Nelson--Aalen estimators of integrated
  transition hazards for non-Markov models}.
\newblock \bibinfo{journal}{\emph{Statistics \& probability letters}}
  \bibinfo{volume}{55}, \bibinfo{number}{4} (\bibinfo{year}{2001}),
  \bibinfo{pages}{403--411}.
\newblock


\bibitem[De~Wreede et~al\mbox{.}(2010)]%
        {de2010mstate}
\bibfield{author}{\bibinfo{person}{Liesbeth~C De~Wreede},
  \bibinfo{person}{Marta Fiocco}, {and} \bibinfo{person}{Hein Putter}.}
  \bibinfo{year}{2010}\natexlab{}.
\newblock \showarticletitle{The mstate package for estimation and prediction in
  non-and semi-parametric multi-state and competing risks models}.
\newblock \bibinfo{journal}{\emph{Computer methods and programs in
  biomedicine}} \bibinfo{volume}{99}, \bibinfo{number}{3}
  (\bibinfo{year}{2010}), \bibinfo{pages}{261--274}.
\newblock


\bibitem[de~Wreede et~al\mbox{.}(2011)]%
        {de2011mstate}
\bibfield{author}{\bibinfo{person}{Liesbeth~C de Wreede},
  \bibinfo{person}{Marta Fiocco}, \bibinfo{person}{Hein Putter},
  {et~al\mbox{.}}} \bibinfo{year}{2011}\natexlab{}.
\newblock \showarticletitle{mstate: an R package for the analysis of competing
  risks and multi-state models}.
\newblock \bibinfo{journal}{\emph{Journal of statistical software}}
  \bibinfo{volume}{38}, \bibinfo{number}{7} (\bibinfo{year}{2011}),
  \bibinfo{pages}{1--30}.
\newblock


\bibitem[Fawcett(2006)]%
        {fawcett2006introduction}
\bibfield{author}{\bibinfo{person}{Tom Fawcett}.}
  \bibinfo{year}{2006}\natexlab{}.
\newblock \showarticletitle{An introduction to ROC analysis}.
\newblock \bibinfo{journal}{\emph{Pattern recognition letters}}
  \bibinfo{volume}{27}, \bibinfo{number}{8} (\bibinfo{year}{2006}),
  \bibinfo{pages}{861--874}.
\newblock


\bibitem[Groha et~al\mbox{.}(2020)]%
        {groha2020neural}
\bibfield{author}{\bibinfo{person}{Stefan Groha}, \bibinfo{person}{Sebastian~M
  Schmon}, {and} \bibinfo{person}{Alexander Gusev}.}
  \bibinfo{year}{2020}\natexlab{}.
\newblock \showarticletitle{Neural ODEs for Multi-State Survival Analysis}.
\newblock \bibinfo{journal}{\emph{arXiv preprint arXiv:2006.04893}}
  (\bibinfo{year}{2020}).
\newblock


\bibitem[Hoff et~al\mbox{.}(2019)]%
        {hoff2019landmark}
\bibfield{author}{\bibinfo{person}{Rune Hoff}, \bibinfo{person}{Hein Putter},
  \bibinfo{person}{Ingrid~Sivesind Mehlum}, {and} \bibinfo{person}{Jon~Michael
  Gran}.} \bibinfo{year}{2019}\natexlab{}.
\newblock \showarticletitle{Landmark estimation of transition probabilities in
  non-Markov multi-state models with covariates}.
\newblock \bibinfo{journal}{\emph{Lifetime data analysis}}
  \bibinfo{volume}{25}, \bibinfo{number}{4} (\bibinfo{year}{2019}),
  \bibinfo{pages}{660--680}.
\newblock


\bibitem[Hougaard(1999)]%
        {hougaard1999multi2}
\bibfield{author}{\bibinfo{person}{Philip Hougaard}.}
  \bibinfo{year}{1999}\natexlab{}.
\newblock \showarticletitle{Multi-state models: a review}.
\newblock \bibinfo{journal}{\emph{Lifetime data analysis}} \bibinfo{volume}{5},
  \bibinfo{number}{3} (\bibinfo{year}{1999}), \bibinfo{pages}{239--264}.
\newblock


\bibitem[Jackson(2016)]%
        {jackson2016flexsurv}
\bibfield{author}{\bibinfo{person}{Christopher~H Jackson}.}
  \bibinfo{year}{2016}\natexlab{}.
\newblock \showarticletitle{flexsurv: a platform for parametric survival
  modeling in R}.
\newblock \bibinfo{journal}{\emph{Journal of statistical software}}
  \bibinfo{volume}{70} (\bibinfo{year}{2016}).
\newblock


\bibitem[Kingma and Ba(2014)]%
        {kingma2014adam}
\bibfield{author}{\bibinfo{person}{Diederik~P Kingma} {and}
  \bibinfo{person}{Jimmy Ba}.} \bibinfo{year}{2014}\natexlab{}.
\newblock \showarticletitle{Adam: A method for stochastic optimization}.
\newblock \bibinfo{journal}{\emph{arXiv preprint arXiv:1412.6980}}
  (\bibinfo{year}{2014}).
\newblock


\bibitem[Meira-Machado et~al\mbox{.}(2009)]%
        {meira2009multi}
\bibfield{author}{\bibinfo{person}{Lu{\'\i}s Meira-Machado},
  \bibinfo{person}{Jacobo de U{\~n}a-{\'A}lvarez}, \bibinfo{person}{Carmen
  Cadarso-Su{\'a}rez}, {and} \bibinfo{person}{Per~K Andersen}.}
  \bibinfo{year}{2009}\natexlab{}.
\newblock \showarticletitle{Multi-state models for the analysis of
  time-to-event data}.
\newblock \bibinfo{journal}{\emph{Statistical methods in medical research}}
  \bibinfo{volume}{18}, \bibinfo{number}{2} (\bibinfo{year}{2009}),
  \bibinfo{pages}{195--222}.
\newblock


\bibitem[Putter and Spitoni(2018)]%
        {putter2018non}
\bibfield{author}{\bibinfo{person}{Hein Putter} {and} \bibinfo{person}{Cristian
  Spitoni}.} \bibinfo{year}{2018}\natexlab{}.
\newblock \showarticletitle{Non-parametric estimation of transition
  probabilities in non-Markov multi-state models: The landmark Aalen--Johansen
  estimator}.
\newblock \bibinfo{journal}{\emph{Statistical methods in medical research}}
  \bibinfo{volume}{27}, \bibinfo{number}{7} (\bibinfo{year}{2018}),
  \bibinfo{pages}{2081--2092}.
\newblock


\bibitem[Rahman et~al\mbox{.}(2021)]%
        {rahman2021deeppseudo}
\bibfield{author}{\bibinfo{person}{Md~Mahmudur Rahman}, \bibinfo{person}{Koji
  Matsuo}, \bibinfo{person}{Shinya Matsuzaki}, {and} \bibinfo{person}{Sanjay
  Purushotham}.} \bibinfo{year}{2021}\natexlab{}.
\newblock \showarticletitle{DeepPseudo: Pseudo Value Based Deep Learning Models
  for Competing Risk Analysis}. In \bibinfo{booktitle}{\emph{Proceedings of the
  AAAI Conference on Artificial Intelligence}}, Vol.~\bibinfo{volume}{35}.
  \bibinfo{pages}{479--487}.
\newblock


\bibitem[Rueda et~al\mbox{.}(2019)]%
        {rueda2019dynamics}
\bibfield{author}{\bibinfo{person}{Oscar~M Rueda},
  \bibinfo{person}{Stephen-John Sammut}, \bibinfo{person}{Jose~A Seoane},
  \bibinfo{person}{Suet-Feung Chin}, \bibinfo{person}{Jennifer~L Caswell-Jin},
  \bibinfo{person}{Maurizio Callari}, \bibinfo{person}{Rajbir Batra},
  \bibinfo{person}{Bernard Pereira}, \bibinfo{person}{Alejandra Bruna},
  \bibinfo{person}{H~Raza Ali}, {et~al\mbox{.}}}
  \bibinfo{year}{2019}\natexlab{}.
\newblock \showarticletitle{Dynamics of breast-cancer relapse reveal
  late-recurring ER-positive genomic subgroups}.
\newblock \bibinfo{journal}{\emph{Nature}} \bibinfo{volume}{567},
  \bibinfo{number}{7748} (\bibinfo{year}{2019}), \bibinfo{pages}{399--404}.
\newblock


\bibitem[Spitoni et~al\mbox{.}(2018)]%
        {spitoni2018prediction}
\bibfield{author}{\bibinfo{person}{Cristian Spitoni}, \bibinfo{person}{Violette
  Lammens}, {and} \bibinfo{person}{Hein Putter}.}
  \bibinfo{year}{2018}\natexlab{}.
\newblock \showarticletitle{Prediction errors for state occupation and
  transition probabilities in multi-state models}.
\newblock \bibinfo{journal}{\emph{Biometrical Journal}} \bibinfo{volume}{60},
  \bibinfo{number}{1} (\bibinfo{year}{2018}), \bibinfo{pages}{34--48}.
\newblock


\bibitem[Titman and Putter(2020)]%
        {titman2020general}
\bibfield{author}{\bibinfo{person}{Andrew~C Titman} {and} \bibinfo{person}{Hein
  Putter}.} \bibinfo{year}{2020}\natexlab{}.
\newblock \showarticletitle{General tests of the Markov property in multi-state
  models}.
\newblock \bibinfo{journal}{\emph{Biostatistics}} (\bibinfo{year}{2020}).
\newblock


\bibitem[Zhao and Feng(2020)]%
        {zhao2020deep}
\bibfield{author}{\bibinfo{person}{Lili Zhao} {and} \bibinfo{person}{Dai
  Feng}.} \bibinfo{year}{2020}\natexlab{}.
\newblock \showarticletitle{Deep neural networks for survival analysis using
  pseudo values}.
\newblock \bibinfo{journal}{\emph{IEEE journal of biomedical and health
  informatics}} \bibinfo{volume}{24}, \bibinfo{number}{11}
  (\bibinfo{year}{2020}), \bibinfo{pages}{3308--3314}.
\newblock


\end{thebibliography}

\appendix

\end{document}